\documentclass[letterpaper, 10pt, conference]{ieeeconf}  

\IEEEoverridecommandlockouts                              

\usepackage{graphics} 
\usepackage{epsfig} 
\usepackage{hyperref}

\usepackage{mathptmx} 
\usepackage{times} 
\usepackage{amsmath} 
\usepackage{amssymb}  
\usepackage{threeparttable}
\usepackage{booktabs}
\usepackage{adjustbox}
\usepackage{siunitx}
\usepackage{subcaption}
\usepackage{graphicx}
\usepackage{tikz}
\usepackage{pgfplots}
\usepackage{balance}

\title{\LARGE \bf CloudGripper: An Open Source Cloud Robotics Testbed for Robotic Manipulation Research, Benchmarking and Data Collection at Scale}

\author{Muhammad Zahid and Florian T. Pokorny%
\thanks{The authors are with the School of Electrical Engineering and Computer Science, KTH Royal Institute of Technology {\tt\small mzmi@kth.se; fpokorny@kth.se}.}}
\begin{document}

\maketitle

\begin{abstract}
    We present CloudGripper, an open source cloud robotics testbed, consisting of a scalable, space and cost-efficient design constructed as a rack of 32 small robot arm work cells. Each robot work cell is fully enclosed and features individual lighting, a low-cost custom 5 degree of freedom Cartesian robot arm with an attached parallel jaw gripper and a dual camera setup for experimentation. The system design is focused on continuous operation and features a 10 Gbit/s network connectivity allowing for high throughput remote-controlled experimentation and data collection for robotic manipulation. CloudGripper furthermore is intended to form a community testbed to study the challenges of large scale machine learning and cloud and edge-computing in the context of robotic manipulation. In this work, we describe the mechanical design of the system, its initial software stack and evaluate the repeatability of motions executed by the proposed robot arm design. A local network API throughput and latency analysis is also provided. CloudGripper-Rope-100, a dataset of more than a hundred hours of randomized rope pushing interactions and approximately 4 million camera images is collected and serves as  a proof of concept demonstrating data collection capabilities. A project website with more information is available at \href{https://cloudgripper.org}{https://cloudgripper.org}.
\end{abstract}

\section{Introduction} 
With recent progress in deep learning models for robotics \cite{perceiver, transgrasp, rt1, metamorph} and the use of large transformer architectures \cite{attention}, the robotics community is increasingly faced with a training data acquisition bottleneck in exploring the full potential of these models. While  applications fields such as computer vision and natural language processing benefited from the availability of large scale image datasets \cite{imagenet2009} and internet-scale text datasets \cite{gpt4}, robotic manipulation research is in comparison still data starved with individual authors typically relying on real-world-executions of only a few hours of robotic experiments. In addition, robotic manipulation work cell and robot setups as well as objects used for research still vary widely between research groups making benchmarking and replication of research challenging. 

\begin{figure}[t]
    \centering
    \includegraphics[width=1\columnwidth]{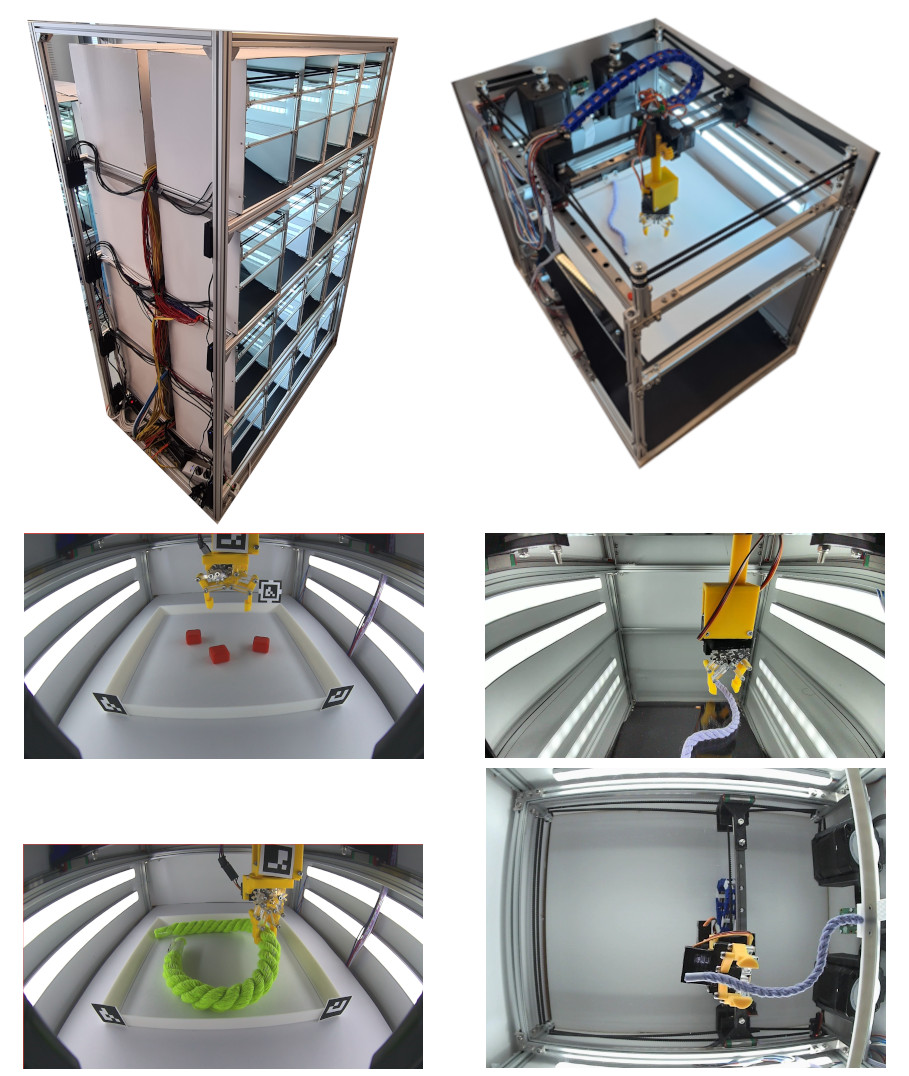}
    \caption{Top left: A CloudGripper rack of 32 robotic arms with front-panels removed for visibility, top right: individual robot cell; middle and bottom left: internal camera images of the system with white floor inlay, QR-marker and 3D-printed cubes and rope manipulation setup; middle and bottom right image: internal top and bottom camera images with transparent base plate to capture multi-view images of  a rope manipulation scenario.}
    \label{fig:rack}
\end{figure}

In this work, we present CloudGripper (see Fig. \ref{fig:rack}), the first rack-based and fully scalable cloud robotics system designed significantly to push the boundaries of parallelism and experimental 
data throughput for robotic manipulation and to serve as a community testbed for large-scale robotic manipulation research and dataset collection at multi terabyte scale. Secondly, CloudGripper offers a first reference implementation of a cloud robotics system of 32 robotic arms using which distributed robotic manipulation and machine learning algorithms can be evaluated and tested.

The core contributions of the present paper are:
\begin{itemize}
\item[1)] We present the system and hardware design of CloudGripper, a first practical and publicly accessible cloud robotics test bed for robotic manipulation with a space-efficient and scalable rack-based design.
\item[2)] We validate our design by means of repeatability tests and provide a throughput analysis based on a single CloudGripper rack of 32 robots.
\item[3)] We present CloudGripper-Rope-100, a dataset consisting of more than a hundred hours of randomized rope manipulations generated using CloudGripper.
\item[4)] We discuss sample use cases of the system and present our vision for an open source community-based development approach of CloudGripper and its software ecosystem with participation of the international community going forward.
\end{itemize}

\section{Related Work} 
\paragraph*{Testbeds in Robotics} The need to improve repeatability of experiments and to find common benchmarking approaches has long been recognized by the robotic manipulation community, with multiple 
datasets, challenges and workshops over the years focussing on this topic \cite{ycb, pickingchallenge, cloudgrasp, cloudcompetition}. The time limited nature of competitions such as the Amazon Picking Challenge \cite{pickingchallenge} and the lack of shared identical experimental setups that comprise not just object datasets such as \cite{ycb} but complete work cells, software stacks and configurations however still inhibits easy replication and comparison of results. In robotics research beyond robotic manipulation more broadly, several testbeds have been proposed to serve as common experimentation platforms \cite{testbeds}. Most well-known perhaps is the Robotarium testbed \cite{robotarium} featuring remote-controlled mobile robots. Similar to our approach, this testbed in particular also opted for low-cost components which ensures the scalability and maintainability of the system. As far as we are aware,
CloudGripper however constitutes the first large scale testbed with a particular focus on remote robotic manipulation at scale. 

\paragraph*{Cloud Robotics} CloudGripper is designed as a reference cloud robotics research platform. The term cloud robotics was coined by James Kuffner in 2010 \cite{kehoe2015survey}, and proposes the use of cloud computing, big data, collaborative learning and crowd sourcing (including open source development) to approach some of the most challenging problems in robotics. Early work focusing on networked and tele-operated robotics, such as the tele-garden project \cite{goldberg2001robot} and pioneering efforts such as the  RoboEarth  research project \cite{waibel2011roboearth} have laid the foundations for cloud robotics and related commercial applications, such as the deployment of autonomous vehicles that are now becoming a reality. Recently, the development of ROS 2.0 \cite{ros2} and efforts such as FogROS2 \cite{fogros2} furthermore continue to push the boundaries of how an edge/cloud approach to robotics can be implemented, but no public robotic manipulation-focused deployments that fully embrace the cloud robotics paradigmn are currently publicly available for research end-users to the best of our knowledge.  

\paragraph*{Scale of Experimentation in Robotics Research} 
State of the art robotic manipulation research labs today require a significant capital, space and staffing investment and are typically equipped with a small number of robotic single and dual arm systems. Systems such as the ABB YuMi \cite{yumi} dual arm robot and the Universal Robots UR5e \cite{ur5e} provide current robotic platform examples in particular.
Since large-scale deployments of such systems are currently beyond the budgets of typical research groups, data collection and experimentation on these systems has also remained relatively small in scale in comparison to web-scale datasets being investigated in large language models \cite{gpt4} for example. Furthermore, data collection remains a labor intensive process even on a single robot. Part of the reason for this is that to ensure safety of both equipment and personnel and due to the need to reset experiments, researchers today still commonly have to physically watch over ongoing experiments on a single robotic system at a time and manually supervise robotic experiments in real-time which is in stark contrast to today's machine learning work-flows, where researchers regularly schedule multiple compute jobs on remote compute clusters in parallel. Due to these obstacles, progress on truly large-scale real-world experimentation in robotic manipulation has been rare and limited to corporate-supported efforts such as Google's pioneering work \cite{levine2018}, Amazon's ARMBench \cite{armbench} and OpenAI's efforts such as \cite{openailearn}. Works that some of the authors of this work also co-authored, such as \cite{moduli} and Dex-Net 1.0 \cite{mahler2016dex} have shown the benefits to generalizability and performance when utilizing larger and larger training datasets - even if based on analytic or simulated scenarios, but scaling laws and exact training data size requirements are still poorly understood. Simulation and sim-to-real transfer \cite{simtoreal} has been proposed as a solution to the lack of large scale real-world training data in particular, but significant challenges remain since simulating complex physical interactions remains challenging. This is in particular the case for the manipulation of deformable objects such as cloth and wires as well as fragile objects. In these cases, simulation still cannot fully match the richness of data collected from real-world robotic experimentation.

\section{System Design} 
CloudGripper is designed as a space-optimized system similar to rack-based servers in a data center. Each rack, as displayed in Fig. \ref{fig:rack}, is constructed from extruded aluminium frames and carries 32 individual robot arms and work cells that are stacked to ensure space-efficiency. In the following, we discuss the mechanical and software design of the system, starting with the design of the integrated individual robot arm and work cell setup used in the system.

\subsection{Integrated Robot Arm Work Cells}
\begin{figure}[htp]
    \centering
    \begin{tikzpicture}
        \node[anchor=south west,inner sep=0] (image) at (0,0) {\includegraphics[width=0.9\columnwidth]{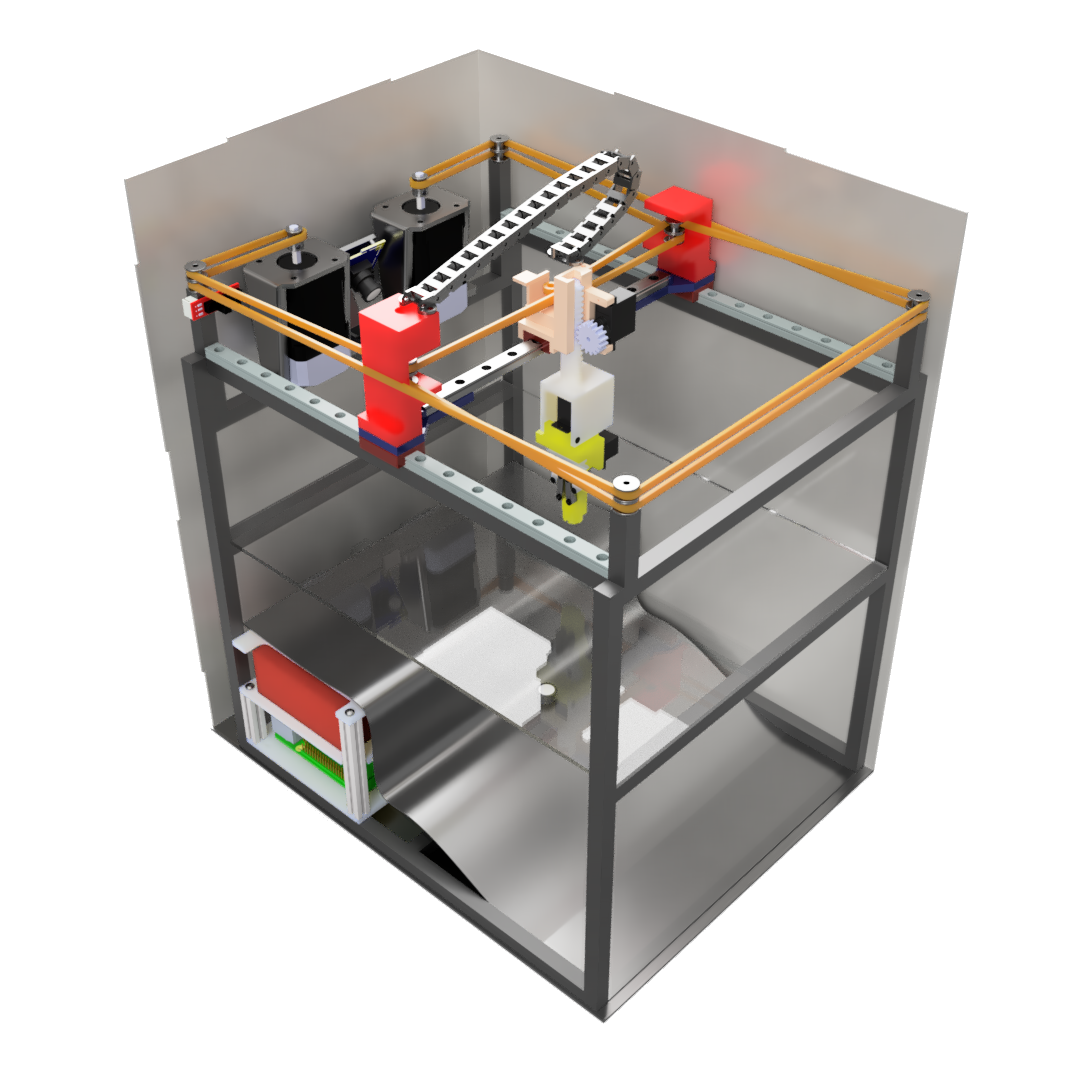}};
        \begin{scope}[x={(image.south east)},y={(image.north west)}]
            \draw[<-,thick] (0.4,0.77) -- (0.4065,0.9677) node[anchor=south] {Stepper Motor};
            \draw[<-,thick] (0.4778, 0.3087) -- (0.3115, 0.1839) node[anchor=east] {Bottom Camera};
            \draw[<-,thick] (0.7689, 0.4519) -- (0.7755, 0.1680) node[anchor=north] {Build Plate};
            \draw[<-,thick] (0.3401, 0.7551) -- (0.1575, 0.9057) node[anchor=east] {Top Camera};
            \draw[<-,thick] (0.5508, 0.6349) -- (0.7, 0.9261) node[anchor=west] {Gripper};
            \draw[<-,thick] (0.2652, 0.3494) -- (0.1211, 0.3586) node[anchor=east] {Electronics};
            \draw[<-,thick] (0.2930, 0.6950) -- (0.1035, 0.5915) node[anchor=east] {CoreXY design};
        \end{scope}
    \end{tikzpicture}
    \caption{Rendering of an individual robot cell}
    \label{fig:robot}
\end{figure}
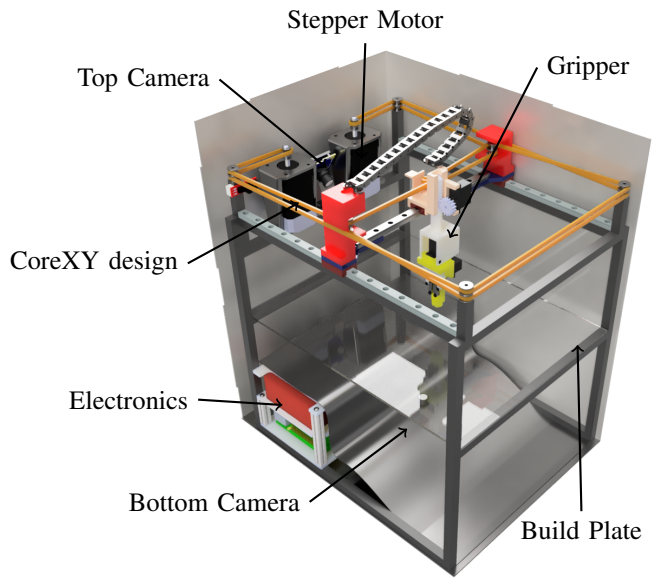
\paragraph*{Work Cell design}
Each CloudGripper robot is housed in a fully enclosed individual work cell featuring 5V and 12V power connections, a 1 Gbit/s Ethernet and a USB-C cable connection. Each work cell has a dimension of $274 \times 356mm\times 400mm$ and features individual led lights and a seal-able front-panel so that no external light is required for operation. For maintenance, the front panel can be opened and custom work cell components, such as 3D-printed parts, for example for a pick-and-place task, can be easily inserted or replaced.

Two RGB cameras are provided in each work cell - a Raspberry Pi 3 wide angle camera providing a top-view of the work environment and a $140$ degree wide-angle camera providing a bottom view through a transparent plexiglass floor plate (see Fig. \ref{fig:rack}). The transparent floor plate is designed for experiments where an additional camera view from below can be beneficial, such as planar pushing tasks. Optionally, a non-transparent inlay can be inserted as shown in the bottom-left of Fig.\ref{fig:rack}, in which case the bottom camera is obstructed and not used. 
The plexi-glass plate also features screw-holes to fixate custom work environments that may be 3D-printed. Each robot cell is constructed from extruded $1\times 1cm$ profile aluminum beams and 3D printed parts. CAD designs will be made available via the project website \href{cloudgripper.org}{cloudgripper.org}. 

\paragraph*{Per-Robot Microcontrollers and Compute}
Each robot features a Raspberry Pi 4B for computation and high-level control which is powered through a USB-C connection. The Raspberry Pi 4B utilizes a Quadcore Cortex-A72 processor with 1.8GHz and is connected to a Teensy 4.1 microcontroller \cite{teensy} based on the ARM Cortex-M7 processor. The microcontroller serves as a low level interface to the robot's sensors and actuators. The Teensy 4.1 architecture was selected due to its ability to provide high frequency control input to the motors in particular. 

\paragraph*{Actuation and Kinematic Structure}

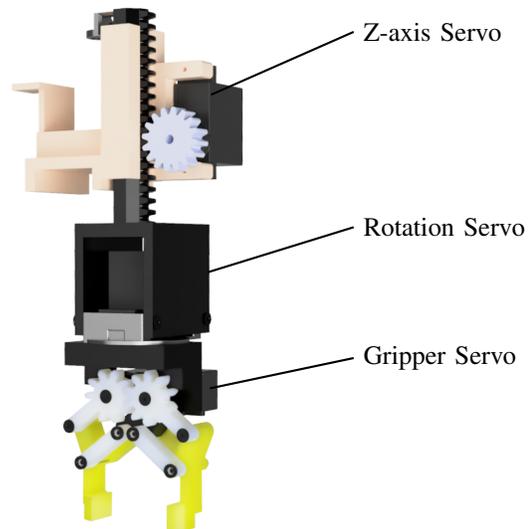
\begin{figure}[t]
    \centering
    \begin{tikzpicture}
        \node[anchor=south west,inner sep=0] (image) at (0,0) {\includegraphics[width=0.68\columnwidth]{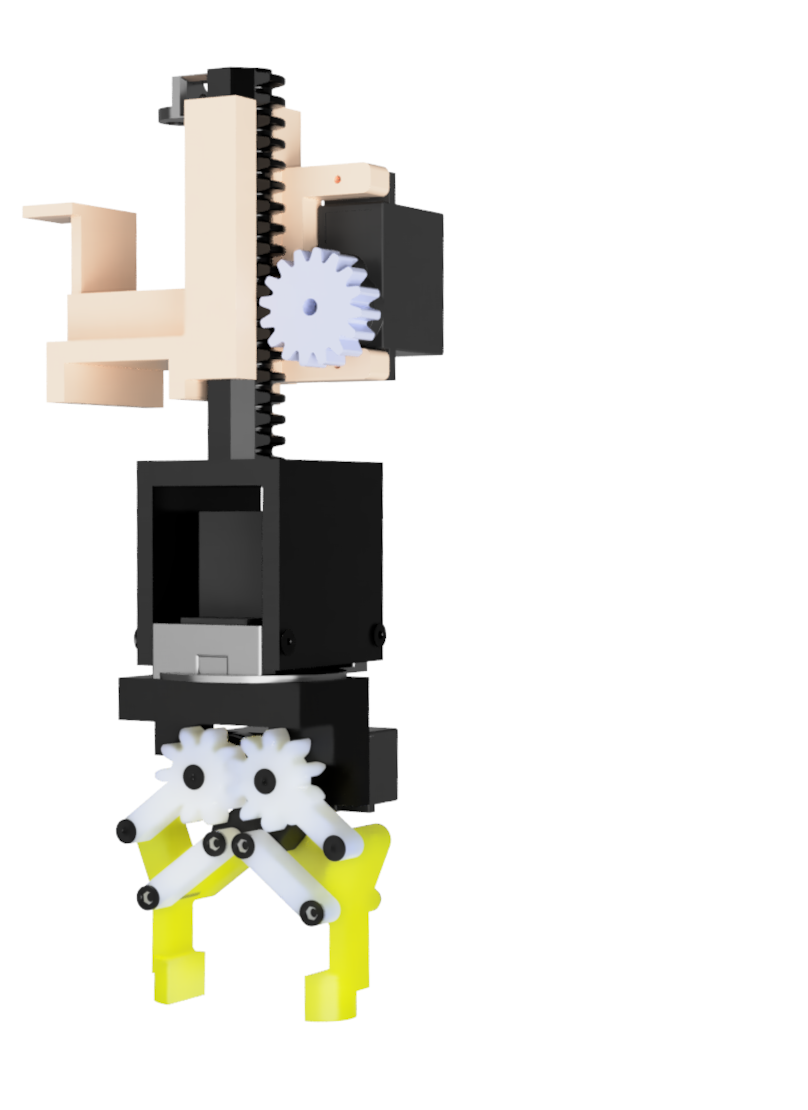}};
        \begin{scope}[x={(image.south east)},y={(image.north west)}]
            \draw[-,thick] (0.4780, 0.7967) -- (0.8, 0.9002) node[anchor=west] {Z-axis Servo};
            \draw[-,thick] (0.4713, 0.5046) -- (0.8, 0.5786) node[anchor=west] {Rotation Servo};
            \draw[-,thick] (0.4755, 0.3133) -- (0.8, 0.3632) node[anchor=west] {Gripper Servo};

        \end{scope}
    \end{tikzpicture}
    \caption{Gripper design with inbuilt servo motors}
    \label{fig:arm}
\end{figure}

Each robot has 5 degrees of freedom $(x, y, z, r, d)$. A Cartesian base that uses two Nema 17 $48 Ncm$ stepper motors with 4000 cpr optical encoders controls the planar $(x, y)$-position using a CoreXY \cite{corexy} motion platform which employs belts as displayed in Fig. \ref{fig:robot}. The CoreXY platform has in particular been used extensively for low-cost 3D-printer designs \cite{corexy} where it has proven its effectiveness. 
The remaining three degrees of freedom (gripper Z-axis $z$, gripper rotation $r$ and parallel jaw gripper opening $d$) are actuated by MG90S micro servos as shown in Fig. \ref{fig:arm}.  A prismatic joint, driven by one of the servo motors accomplishes the z-axis motion in particular while a revolute joint is used to rotate the gripper around its vertical axis and the gripper is opened and closed using a further servo motor as shown in Fig.\ref{fig:arm}. The main parts of the gripper are 3D-printed from PETG while the gears are lasercut from plexiglas. The low-level controller is implemented on the Teensy 4.1 microcontroller. The current implementation allows arbitrary piecewise linear movements in XY direction as well as separate rotation, z motions and gripper opening/closing and employs a preset velocity profile along a given trajectory. Calibration switches are added to re-calibrate XY robot motions.

\paragraph*{Servo actuation and collision detection}

Micro servos provide a suitable choice for the chosen small form factor gripper, but require calibration to ensure uniform performance. Using reference positions from a potentiometer, each servo was therefore calibrated to achieve consistent performance. Since the chosen servos operate via pulse width signals and do not offer feedback on their position, we furthermore integrated current sensors allowing for the detection of collisions. This strategy was used to enable the prevention of motor overheating and potential damage to robot.

\subsection{Software-Stack and Component Communication}
\begin{figure}[t]
    \centering
    \includegraphics[width=1.0\columnwidth]{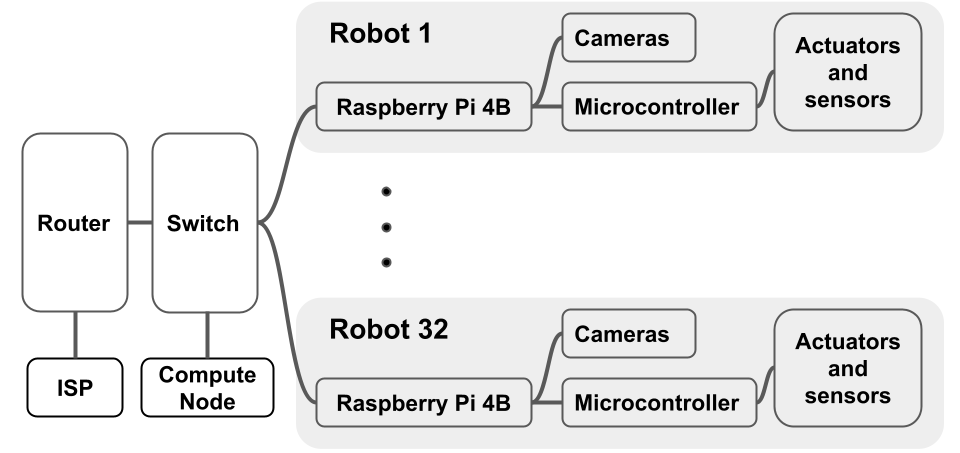}
    \caption{CloudGripper rack networking with 32 robots connected to a switch which enables further communication to the internet via a router as well as access to a local area network server for local computation.}
    \label{fig:architecture}
\end{figure}

CloudGripper's current software stack is focused on simplicity and will be extended significantly in future work. In its current form, each robot's Raspberry Pi 4 runs Raspberry Pi OS and administrators can access each robot via SSH for maintenance. Furthermore, a lean software stack was developed for the initial version of CloudGripper presented here. Each Raspberry runs a REST API server with token-based authentication that is implemented using the python Flask library \cite{flask} and allows remote end-users to request information about a robot's state, to obtain camera images at up to 30fps and to send motion commands that are then forwarded by the API to the microcontroller for execution. Additionally, remote end-users can request livestreams to view the robot in operation. The use of a standard REST API approach proposed here also enables simple development of applications such as teleoperation web-interfaces with buttons for remote control.  Each of the 32 robots in a given CloudGripper rack features an individual 1 Gbit/s Ethernet cable connection to a 10 Gbit/s switch that is then connected to both a router with a 10 Gbit/s internet connection and a 10 Gbit/s connection to a local storage and compute node, see Fig.\ref{fig:architecture} for details on CloudGripper's networking layout.

\subsection{CloudGripper as an Open Source Ecosystem}
CloudGripper is released as an Open Source project with hardware and software released under permissive licensing. Beyond software and hardware, it is however our hope that the project will furthermore evolve into a living testbed with many international users where novel algorithms can be tested against open reference-implementations. 
The project website \href{cloudgripper.org}{cloudgripper.org} will serve as a central resource as the project develops and we intend to 
also host a growing selection of datasets there, starting 
with the CloudGripper-Rope-100 dataset which is described in this paper.


\section{System Design Validation} 
To validate the design of CloudGripper, we present a sequence of experimental evaluations testing both individual robot repeatability and behaviour under extended load.

\subsection{Repeatability analysis}
\begin{figure}[htp]
    \centering
    \input{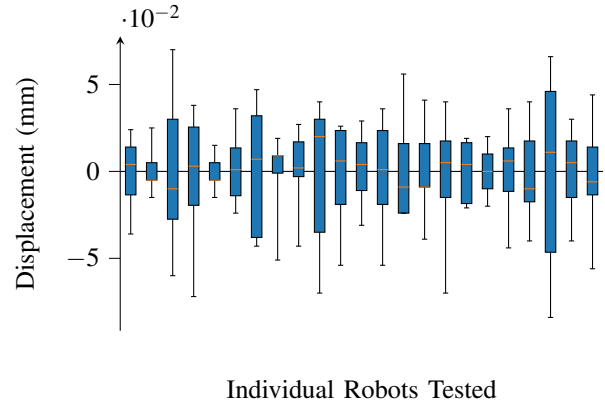}
    \caption{Box-plots of repeatability measurements in XY axis for 23 CloudGripper robots.
    Observed measurement ranges are displayed together with lower and upper quartiles which are indicated by blue boxes. The median of the measurements for each robot is shown as an orange line.}
    \label{fig:xy_repeatability}
\end{figure}
\begin{figure}[htp]
    \centering
    \input{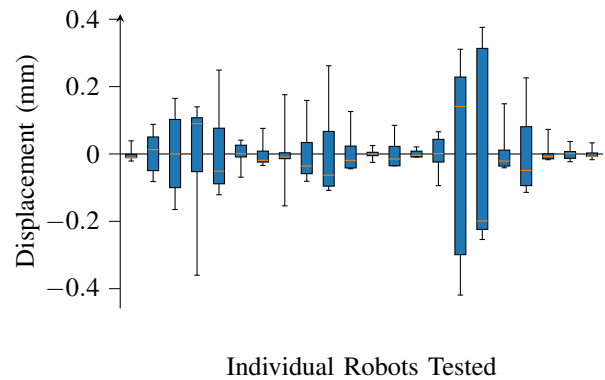}
    \caption{Box-plots of repeatability measurements in Z axis for 23 CloudGripper robots.
    Observed measurement ranges are displayed together with lower and upper quartiles which are indicated by blue boxes. The median of the measurements for each robot is shown as an orange line.
    }
    \label{fig:z_repeatability}
\end{figure}
For this evaluation, the repeatability of the robots' motions was measured on a sample of 23 robots in a CloudGripper rack using a digital dial indicator in a fixed sample location for Z-axis measurements, representing the vertical displacement set by the z-axis servo motor. Furthermore XY-axis measurements, capturing the lateral displacement caused by the stepper motors were collected. For each, the robot was moved linearly between a set of 5 uniformly sampled waypoints in Z and XY dimension respectively before returning to its initial configuration where data from the dial indicator was recorded. This test was performed 10 times for each sample location and robot. 

The distribution of XY-axis measurements is shown in Figure \ref{fig:xy_repeatability} while the distribution of Z-axis measurements is displayed in Figure \ref{fig:z_repeatability}. 
For the XY-axis measurements, the range of deviation in position was between $-0.084 mm$ and $0.070 mm$ and the standard deviation across all tests was calculated to be around $0.027 mm$, highlighting the consistency of the robot's XY-axis movements. For the Z-axis measurements, the range of deviation was higher and between $-0.419 mm$ and  $0.376 mm$ with a standard deviation of approximately $0.109 mm$. While still relatively minor in absolute scale given the low cost nature of the system components, some of the tested robots exhibited variability in repeatability in Z-axis measurements in particular. We hypothesize that variability of the servo motors used in these robots as well as variations in the 3D-printed linear z-axis gear may be contributing factors to these differences.

\subsection{Network Stress-Test}
\begin{figure}[htp]
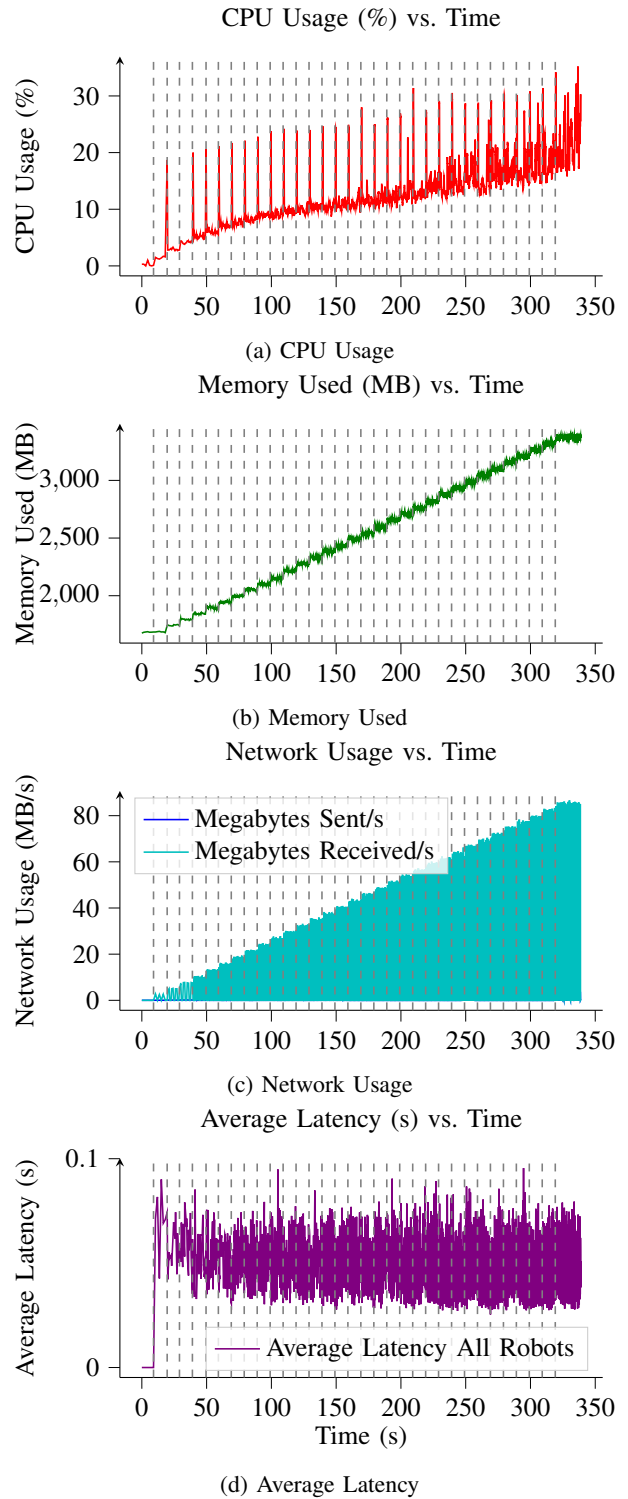

    \centering
    
    \begin{subfigure}{\columnwidth}
        \centering
        \input{cpu_usage.tex}
        \caption{CPU Usage}
    \end{subfigure}
    
    \begin{subfigure}{\columnwidth}
        \centering
        \input{memory_used.tex}
        \caption{Memory Used}
    \end{subfigure}
    
    \begin{subfigure}{\columnwidth}
        \centering
        \input{network_usage.tex}
        \caption{Network Usage}
    \end{subfigure}
    
    \begin{subfigure}{\columnwidth}
        \centering
        \input{average_latency.tex}
        \caption{Average Latency}
    \end{subfigure}
    
    \caption{CloudGripper network stress test evaluation conducted by sending concurrent API camera image requests. Here the number of concurrent robots queried increases at each vertical dashed line by one starting with one robot on the left and terminating with all 32 robots in the rightmost part of the plot.
    We observe near-linear increases in CPU and memory usage on the querying server while API request latency remains stable and only a small fraction of the total network bandwidth of 10 Gbit/s is utilized even when all robots are queried simultaneously.}
    \label{fig:stress}
\end{figure}

One of the core objectives of our design is to facilitate scalability of experimentation by allowing users to  control multiple robots simultaneously via API requests. We performed a network stress test where the local compute server, connected via LAN to CloudGripper, sent API requests for still images from both cameras of a given robot at 10 fps. Starting with a single robot, we incrementally add an additional robot to these concurrent queries with one new thread per query process until all 32 robots are receiving concurrent api requests. As shown in Fig.\ref{fig:stress}, CPU and memory utilization on the querying server steadily increase with each new query thread. 

As expected, network bandwidth usage increases linearly to approximately 80 megabytes per second, utilizing only less than a tenth of the theoretical maximal network bandwidth of 10 Gbit/s and thus illustrating the potential of the system to handle high concurrent loads. 

Conversely, this result also indicates that a 10 Gbit/s network is likely able to accommodate a significantly larger number of robots. 

Additionally, the maximal latency for all requests and robots remains below 100ms during these tests with a 95th percentile latency of approximately 70ms. We note that these performance metrics can likely be improved significantly in future since both the API server and requesting code used during these tests are based on simple python scripts written without any particular focus on optimization.

\section{CloudGripper-Rope-100 Dataset}
To demonstrate data collection capabilities and to stress-test the system with continuous use, we collected a dataset of randomized rope manipulations on 19 robots that were available for data collection. 

Rope manipulation was selected as an example of a challenging manipulation problem due the inherent complexities of state estimation and prediction when manipulating deformable objects \cite{modelingdeform}. 

\begin{figure}[htb]
\centering
\includegraphics[width=0.85\columnwidth]{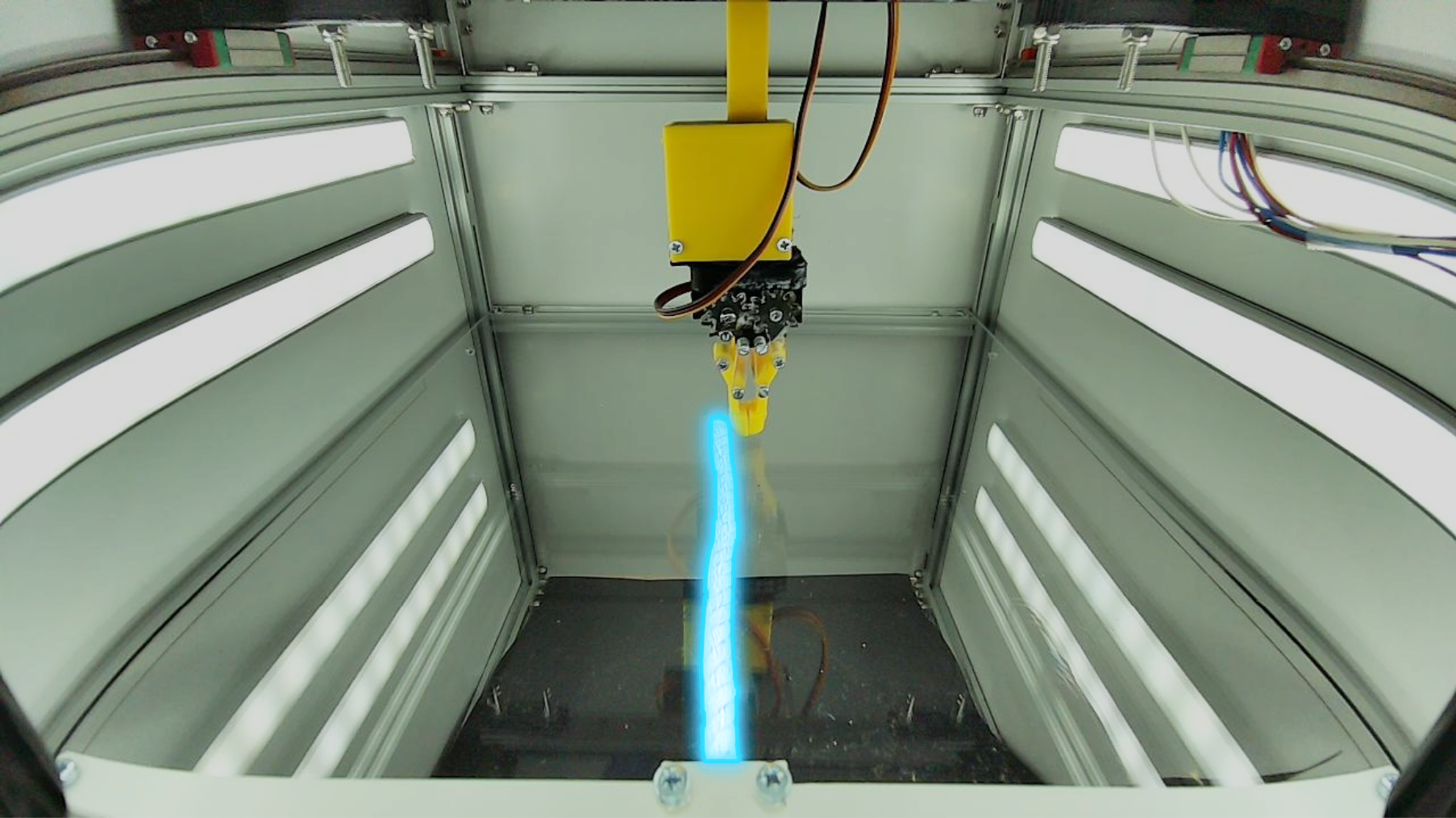}
\\ \vspace{2mm}
\includegraphics[width=0.85\columnwidth]{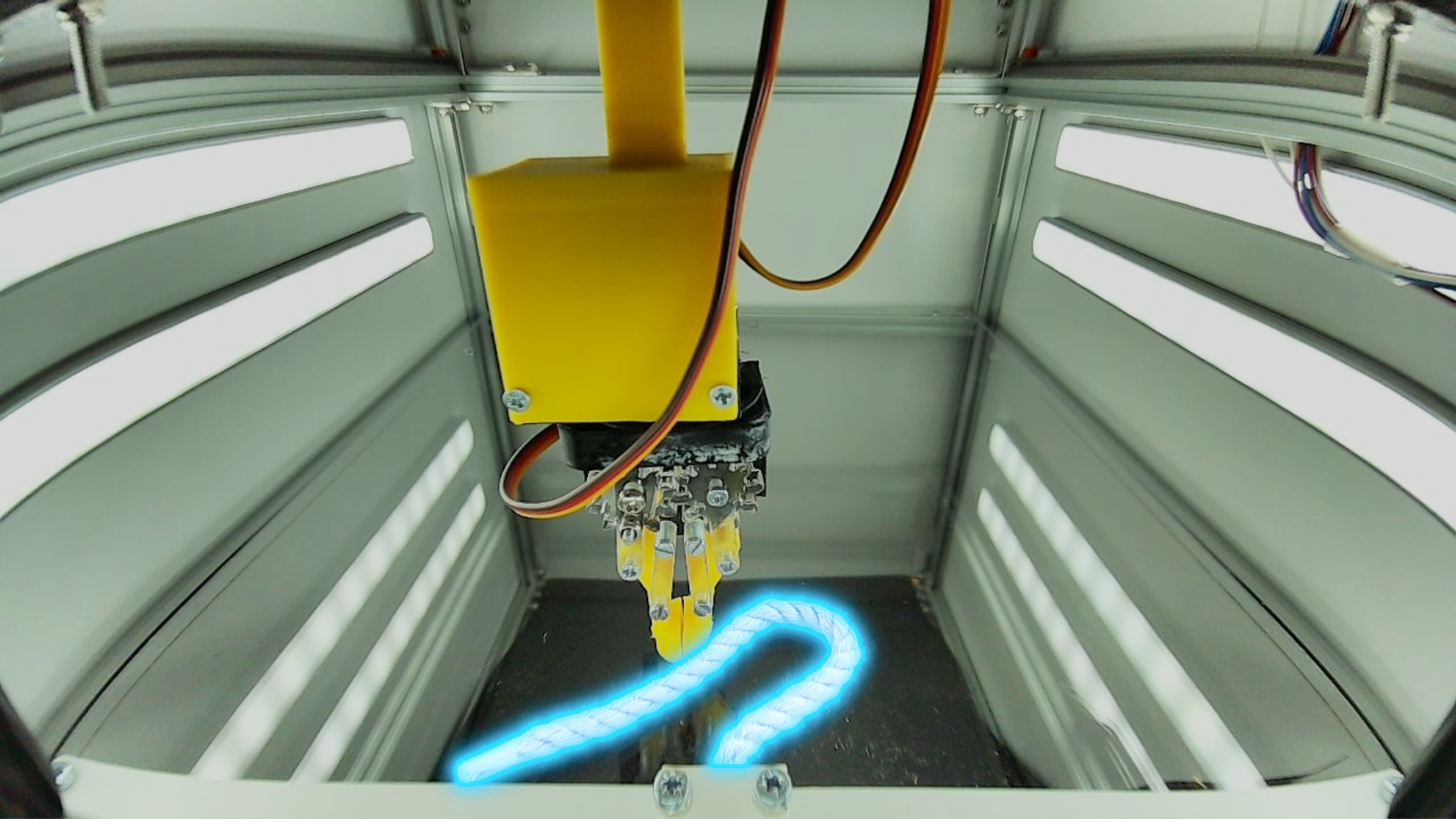}
\\ \vspace{2mm}
\includegraphics[width=0.42\columnwidth]{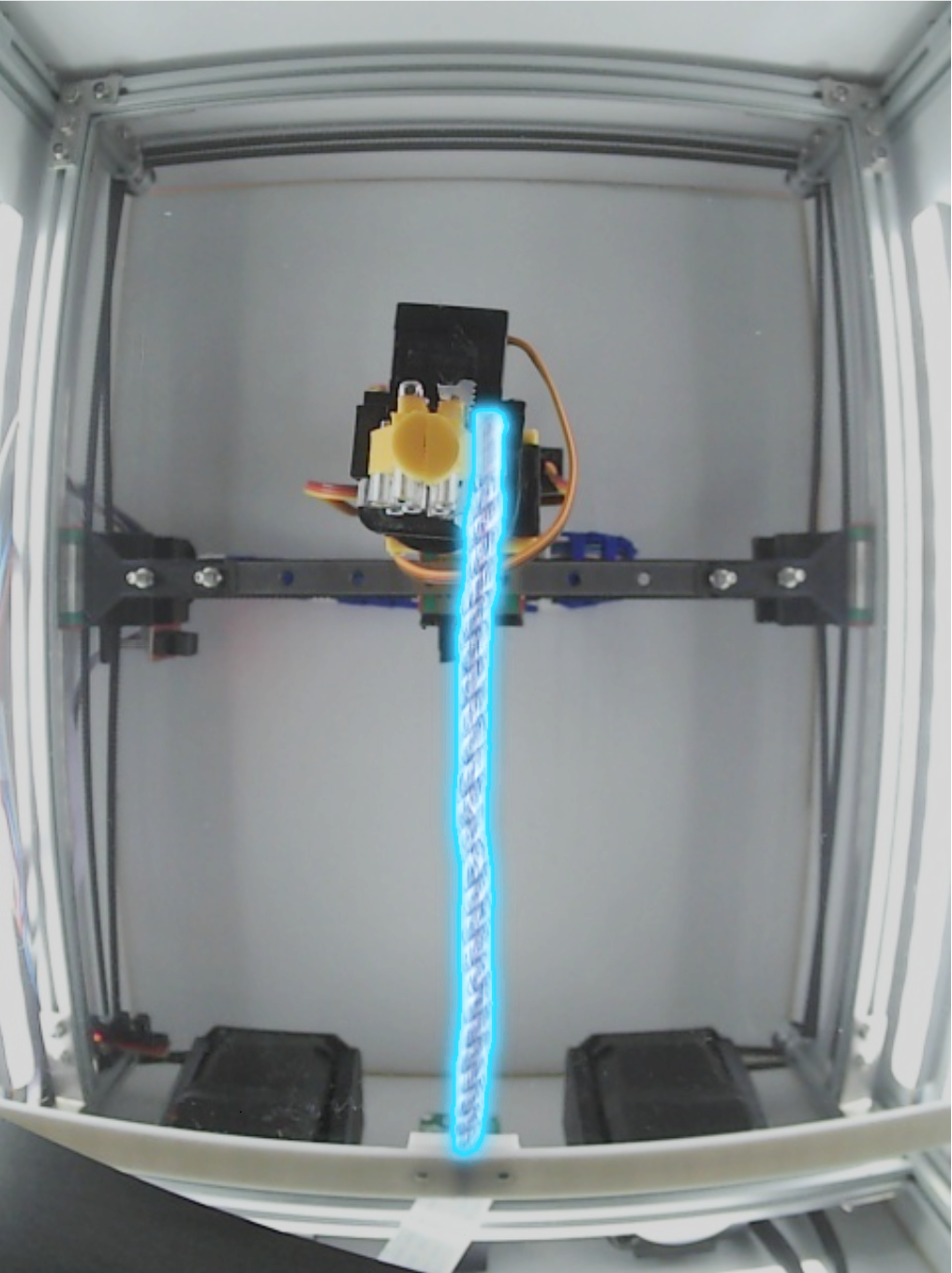}\hspace{2mm}%
\includegraphics[width=0.42\columnwidth]{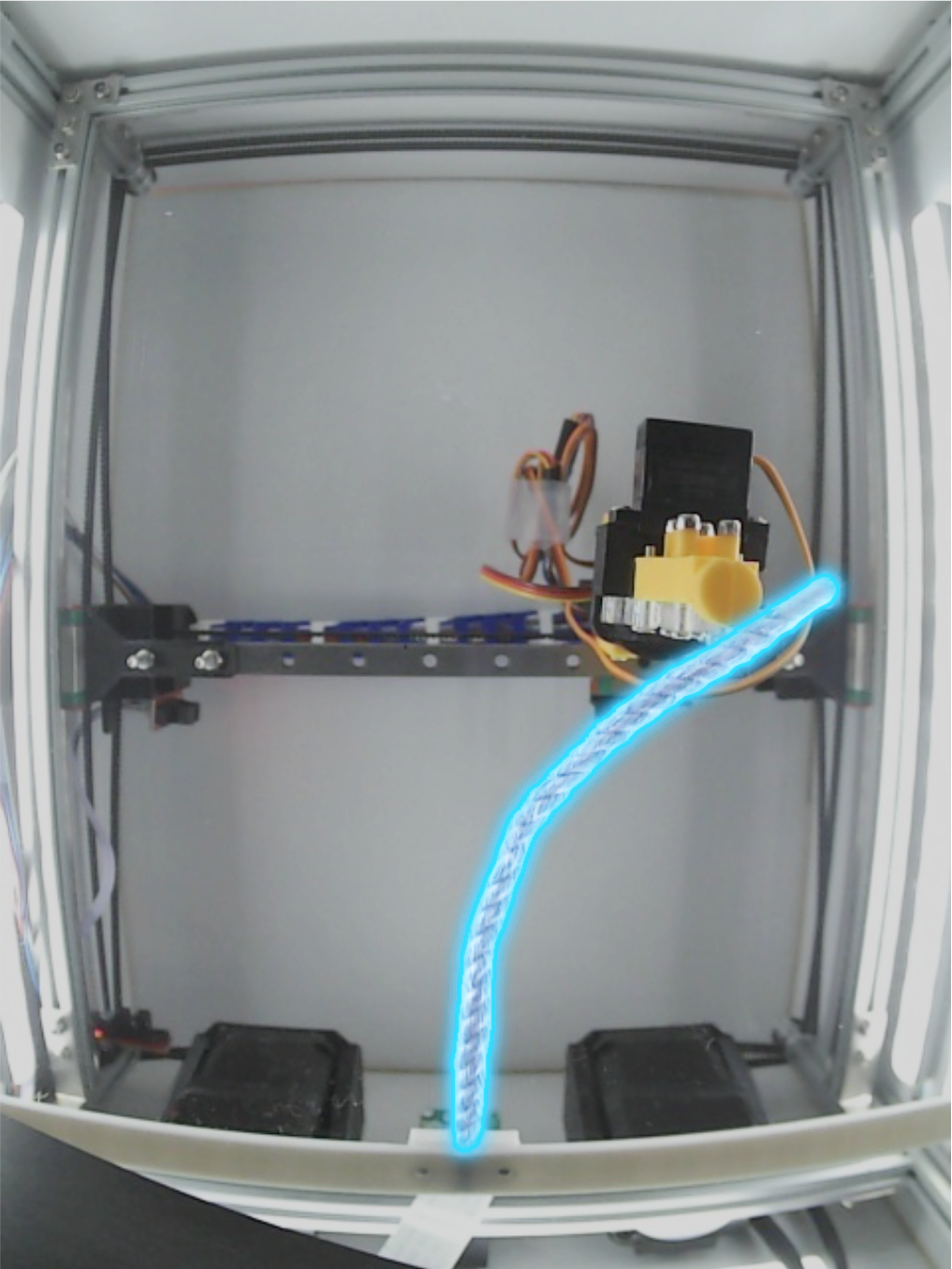}%
\\ \vspace{2mm}
\includegraphics[width=0.42\columnwidth]{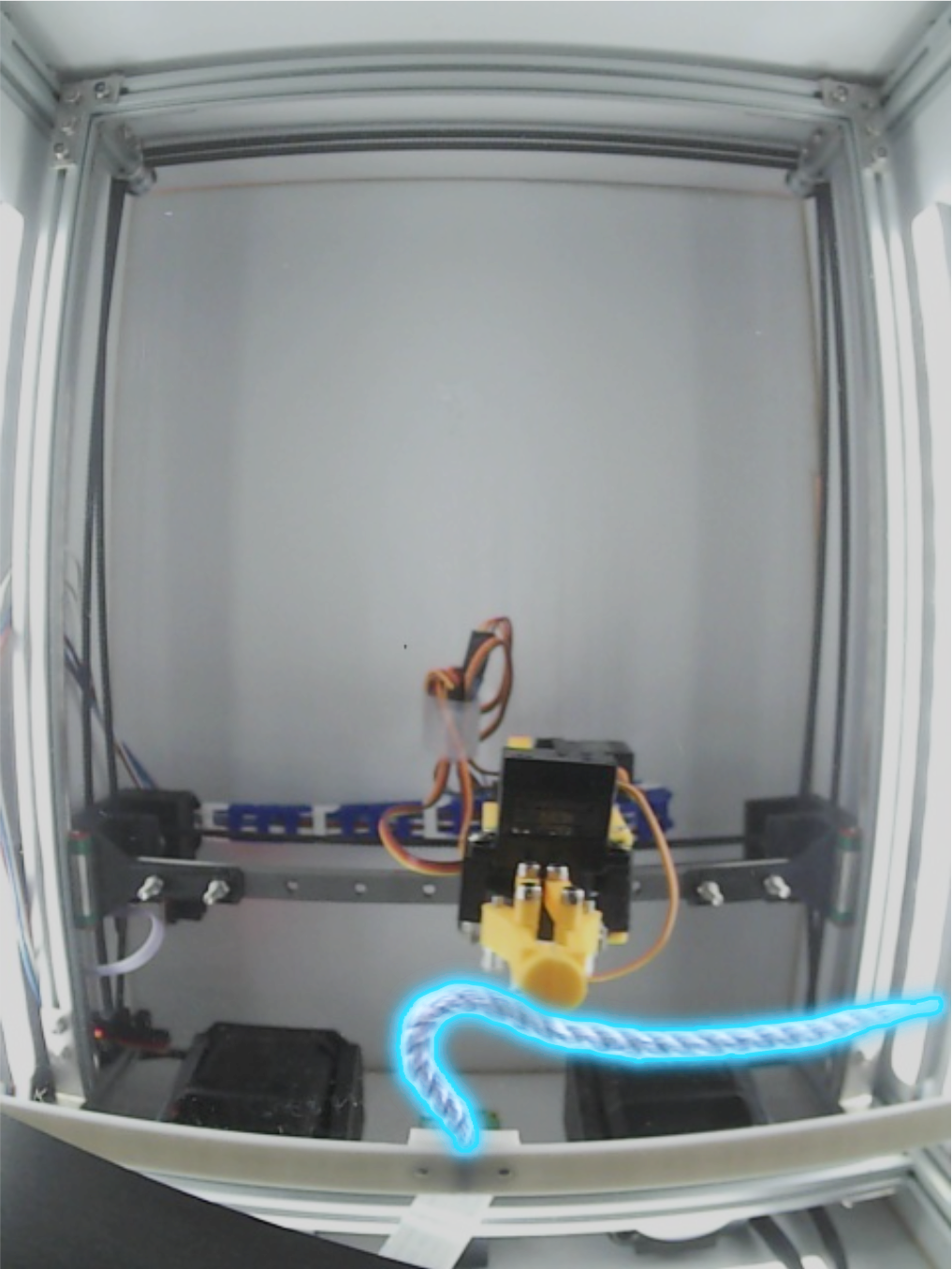}\hspace{2mm}%
\includegraphics[width=0.42\columnwidth]{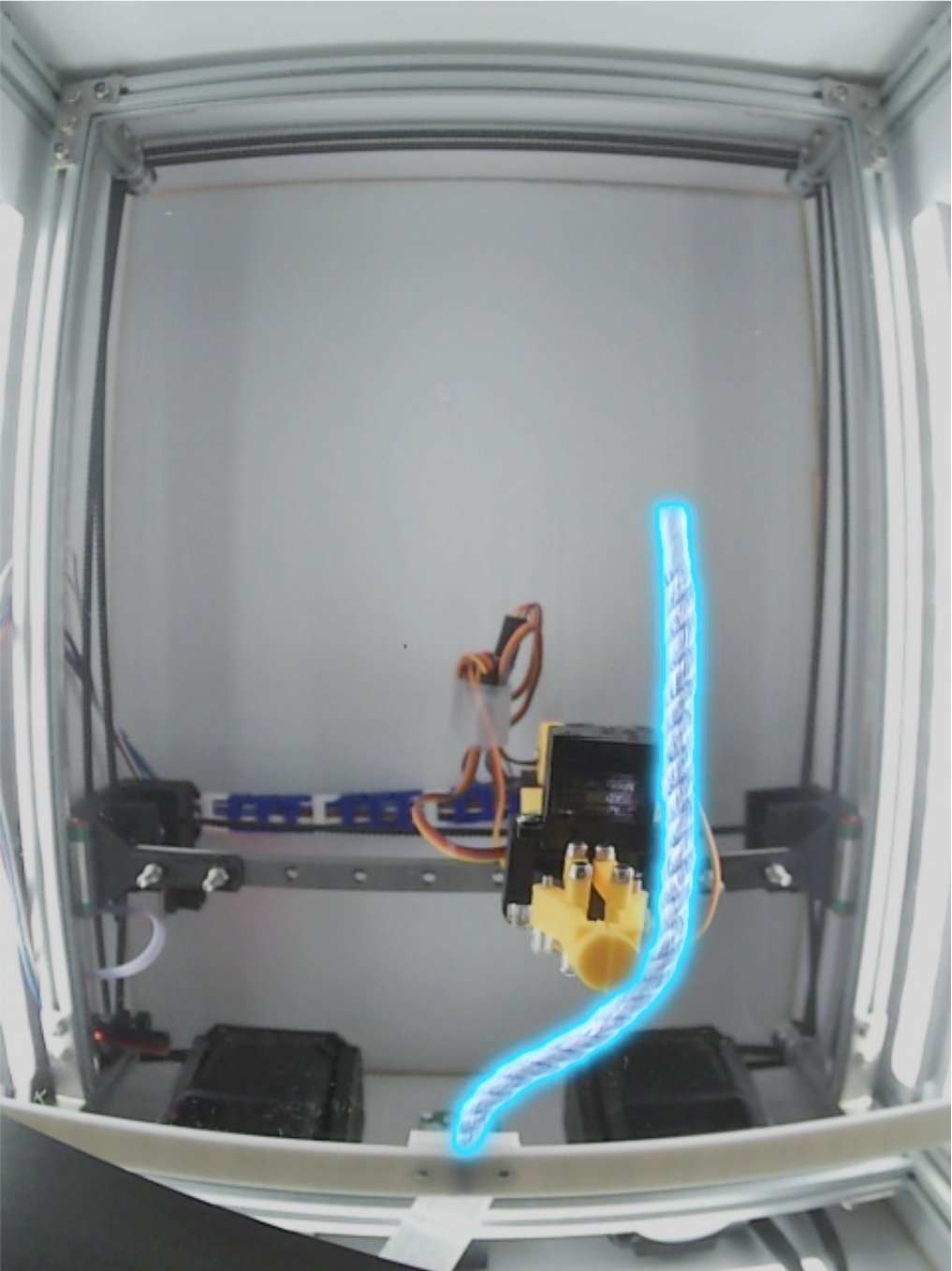}%

\caption{Images from the CloudGripper-Rope-100 manipulation dataset. The top two rows display sample top camera views while bottom rows display bottom camera views. In all cases a rope segmentation using Segment Anything \cite{segmentanything} is applied to highlight the rope in the figures.}
\label{fig:rope}
\end{figure}

To collect data, 19 robots were fitted with a custom work environment that featured a rope affixed to one side of the workspace as seen in Fig. \ref{fig:rope}.
The robot gripper is lowered to the glass plate and remains in this $z$-position during the data collection throughout. Only the $xy$-coordinate is modified to execute pushing of the rope.

Data generation is initialized with the rope in a straightened position as seen in the top left of Fig. \ref{fig:rope}.
On each robot candidate planar linear robot pushing motions are sampled uniformly. Using a simple color segmentation of the rope, we reject candidate linear $xy$-motion trajectories that do not result in a predicted collision with the rope. Candidate motions that result in intersection are executed, resulting in up to 10 consecutive random perturbations of the rope by linear pushing actions per episode. The rope is approximately reset using an automated resetting procedure that attempts to straighten the rope again. During the rope pushing process, camera images are captured at 10 fps and linear pushing actions are executed at a frequency of approximately 0.5Hz. 

The resulting CloudGripper-Rope-100 dataset resulted in approximately 110 hours of robot motion data and approximately 4 million individual camera images of pushing episodes. 

The combined dataset requires approximately half a terabyte of storage space and contains the following components:
\begin{itemize}
    \item \textbf{Metadata}: Robot identifiers, description of the robot environment, total duration of data collection, dataset size, cumulative number motion commands sent and  camera calibration matrices are provided.
    \item \textbf{Robot Actions:} The time stamped API motion commands for each robot are stored to allow for replay of episodes on both real and simulated environments.
    \item \textbf{Camera Images:} Both camera views are stored as color images at 10 fps per robot with a resolution of
    $1280\times720$ pixels for the top camera and $640\times 480$ pixels for the bottom camera. 
\end{itemize}

\section{Discussion of Future Use Cases}
\paragraph*{Training Datasets} 
The generation of rich large scale training datasets is an obvious use case for CloudGripper that we have started to demonstrate in the presented rope manipulation dataset. Future datasets focused on specific object types or task 
settings including cloth folding, pick-and-place tasks as well as simplified assembly tasks could be considered. The collection of human demonstration data by means of teleoperation by multiple remote users could also provide a valuable direction.
\paragraph*{Iterative Algorithm Development} CloudGripper presents an opportunity for researchers to test robotics algorithms in a real-world context. A key question going forward will be to identify robot workspace setups and interaction scenarios that are of most interest to potential collaborators.
\paragraph*{Benchmarking} By offering a standardized robot environment, CloudGripper may be used as a benchmarking and reference implementation platform on which common perception, planning and manipulation algorithms can be compared.
\paragraph*{Transfer Learning} The availability of several near-identical robot cells offers a research opportunity to reason about transfer learning  between tasks and physical workspace setups in a parallelized manner.

\section{System Limitations and Future Work}
The CloudGripper system proposed here provides a first reference design of a large scale robotic manipulation cloud robotics research platform. The design is optimized for cost-efficiency, rapid prototyping and scalability instead of featuring more complex 6 or 7 degree of freedom robot arms made from precision machined components. As a result, the current robotic arm system is of course limited in the type of manipulations it can perform. We believe however that even at its current form, the unique scale offered by this approach will unlock novel insights in data-driven robotics research that were simply not accessible previously. A further aspect is that CloudGripper enables researchers to test novel cloud and edge computing strategies for robotics which we shall explore further in future. 
We also intend to provide
ROS2 \cite{ros2} wrappers and a simulation environment for Sim-To-Real experimentation in the near future.

\section{Acknowledgements}
 This work was partially supported by the Wallenberg AI, Autonomous Systems and Software Program (WASP) funded by the Knut and Alice Wallenberg Foundation.
\bibliographystyle{IEEEtran}
\balance
\bibliography{citations}

\end{document}